\documentclass[sigconf]{acmart}
\usepackage[utf8]{inputenc} 
\usepackage[T1]{fontenc}    
\usepackage{hyperref}       
\usepackage{amsfonts}       
\usepackage{amsmath}

\usepackage{amssymb}
\usepackage{amsthm}
\usepackage[capitalize,noabbrev]{cleveref}
\usepackage{bm}
\usepackage{algorithm}
\usepackage{algorithmic}
\usepackage{graphicx}
\usepackage{subcaption}
\usepackage{pifont}
\usepackage{adjustbox}
\usepackage{wrapfig}
\usepackage{balance}
\newcommand{\xmark}{\ding{55}}%

\theoremstyle{plain}

\theoremstyle{definition}

\theoremstyle{remark}

\AtBeginDocument{%
  }

\copyrightyear{2025}
\acmYear{2025}
\setcopyright{acmlicensed}\acmConference[KDD '25]{Proceedings of the 31st ACM SIGKDD Conference on Knowledge Discovery and Data Mining V.1}{August 3--7, 2025}{Toronto, ON, Canada}
\acmBooktitle{Proceedings of the 31st ACM SIGKDD Conference on Knowledge Discovery and Data Mining V.1 (KDD '25), August 3--7, 2025, Toronto, ON, Canada}
\acmDOI{10.1145/3690624.3709341}
\acmISBN{979-8-4007-1245-6/25/08}

\begin{document}

\title{Task Diversity in Bayesian Federated Learning: Simultaneous Processing of Classification and Regression}


\author{Junliang Lyu}
\affiliation{%
\institution{Center for Applied Statistics and School of Statistics, Renmin University of China}
\city{Beijing}
\country{China}}
\email{lvjunliang0211@ruc.edu.cn}

\author{Yixuan Zhang}
\affiliation{%
  \institution{School of Statistics and Data Science, \\Southeast University}
  \city{Nanjing}
  \country{China}
}
\email{zh1xuan@hotmail.com}

\author{Xiaoling Lu}
\affiliation{%
 \institution{Center for Applied Statistics and School of Statistics, Renmin University of China}
  \institution{Innovation Platform, Renmin University of China}
 \city{Beijing}
 \country{China}
}
\email{xiaolinglu@ruc.edu.cn}

\author{Feng Zhou}
\authornote{Corresponding Author}
\affiliation{%
  \institution{Center for Applied Statistics and School of Statistics, Renmin University of China}
  \institution{Beijing Advanced Innovation Center for Future Blockchain and Privacy Computing}
  \city{Beijing}
  \country{China}
}
\email{feng.zhou@ruc.edu.cn}

\renewcommand{\shortauthors}{Junliang Lyu, Yixuan Zhang, Xiaoling Lu and Feng Zhou}

\begin{abstract}
    This work addresses a key limitation in current federated learning approaches, which predominantly focus on homogeneous tasks, neglecting the task diversity on local devices. 
    We propose a principled integration of multi-task learning using multi-output Gaussian processes (MOGP) at the local level and federated learning at the global level. 
    MOGP handles correlated classification and regression tasks, offering a Bayesian non-parametric approach that naturally quantifies uncertainty. 
    The central server aggregates the posteriors from local devices, updating a global MOGP prior redistributed for training local models until convergence. 
    Challenges in performing posterior inference on local devices are addressed through the P\'{o}lya-Gamma augmentation technique and mean-field variational inference, enhancing computational efficiency and convergence rate. 
    Experimental results on both synthetic and real data demonstrate superior predictive performance, OOD detection, uncertainty calibration and convergence rate, highlighting the method's potential in diverse applications. Our code is publicly available at \href{https://github.com/JunliangLv/task_diversity_BFL}{\textcolor{blue}{https://github.com/JunliangLv/task\_diversity\_BFL}}.
\end{abstract}



\received{8 August 2024}
\received[revised]{18 October 2024}
\received[accepted]{16 November 2024}

\maketitle

\section{Introduction}
\label{introduction}
Over the past few years, artificial intelligence has experienced tremendous growth. Traditional machine learning methods often necessitated centralizing datasets for training. However, with the proliferation of edge devices like smartphones and Internet of Things (IoT) devices, there is a strong demand for machine learning models to be trained on dispersed data. Therefore, federated learning (FL)~\cite{yang2019federated} has emerged as a concept in recent years, aiming to train models using data scattered across multiple local devices, thus avoiding large-scale data transfers and enhancing data privacy~\cite{zhang2021survey}. 

While FL has seen considerable advancement, it is known that most current FL efforts focus on homogeneous tasks on local devices, either exclusively for classification or solely for regression tasks. However, this contradicts real-world scenarios, where local devices often gather data for both types of tasks. 
Taking the health monitoring application on a smartphone as an example: it collects various health metrics such as heart rate, step count, and sleep quality. 
Suppose the application aims to classify the user's movement states, such as stationary or walking, using sensor data like step count. Simultaneously, it can utilize heart rate and sleep duration for regression analysis, predicting trends in specific indicators. 
It is evident that this example involves both classification and regression tasks, and they are closely correlated. This implies a need to adopt multi-task learning (MTL) approaches to simultaneously handle both types of tasks on the local device. 

Furthermore, numerous existing FL frameworks rely on deterministic methods, suffering from overfitting when data is limited and providing predictions without uncertainty estimation, restricting their application in high-risk domains. 
For example, in high-risk domains, encountering decisions with high uncertainty indicates a need for caution, prompting a shift towards conservative strategies rather than complete reliance on algorithmic outputs. Similarly, in the context of out-of-distribution (OOD) detection, leveraging uncertainty helps identify OOD samples as they tend to exhibit higher uncertainty compared to in-distribution data. 

This article aims to integrate multi-task learning at the local level and federated learning at the global level in a principled probabilistic manner. Specifically, on each local device, we employ \emph{multi-output Gaussian processes} (MOGP)~\cite{alvarezRL12kernel} to jointly model multiple correlated classification and regression tasks. As a Bayesian framework, MOGP naturally quantifies uncertainty through posterior inference. 
On the central server, we aggregate the posteriors uploaded by local devices to obtain an updated global MOGP prior. This updated global prior is then redistributed to local devices to train new local models. 
This iteration continues until the global convergence. 

It is worth noting that performing posterior inference on local devices presents a challenge due to the non-conjugacy of classification likelihood with MOGP prior, requiring approximation methods like Markov chain Monte Carlo (MCMC)~\cite{neal1993probabilistic} or variational inference (VI)~\cite{blei2017variational}. While VI is computationally efficient, the standard methods, which assume a Gaussian variational distribution and optimize a tractable evidence lower bound (ELBO), often suffer from slow convergence~\cite{wenzel2019efficient}. 
To address this challenge, this work employs the P\'{o}lya-Gamma augmentation technique~\cite{polson2013bayesian}, crafting a mean-field VI with closed-form expressions.

Specifically, we make the following contributions: 
\textbf{(1)} at the local level, we extend from single-task to multi-task settings, empowering local device to handle correlated classification and regression tasks concurrently; 
\textbf{(2)} as a Bayesian approach, our local model not only provides predictions but also characterizes uncertainty, a crucial factor in OOD detection and model calibration; 
\textbf{(3)} by enhancing local posterior inference using P\'{o}lya-Gamma augmentation, we derive a completely analytical mean-field VI method, significantly boosting convergence; 
\textbf{(4)} across synthetic and real datasets, our method outperforms baselines in predictive performance, and demonstrates superior OOD detection, uncertainty calibration, and fast convergence. 
Lastly, we conduct ablation studies to explore the robustness of our method concerning various components.

\section{Related Works}
\label{related work}
In this section, we discuss pertinent research on FL, Bayesian FL, and multi-task learning. 

\subsection{Federated Learning}
In FL, collaboration among clients is pivotal for addressing learning tasks while upholding data privacy. 
Google introduced the initial FL algorithm, FedAvg, to safeguard client privacy in distributed learning~\cite{mcmahan2017communication}.
Subsequent advancements encompass a range of methods to enhance convergence~\cite{li2018convergence,stich2018local,haddadpour2019convergence}, fortify data privacy~\cite{agarwal2018cpsgd,truex2019hybrid,wei2020federated}, and improve communication efficiency~\cite{chen2021communication,sattler2019robust,reisizadeh2020fedpaq}. 
Personalized federated learning (PFL) has gained traction in recent years, overcoming the suboptimal performance of early FL methods when confronted with heterogeneous datasets~\cite{rothchild2020fetchsgd}. 
Recent methods include local customization~\cite{t2020personalized,huang2021personalized,tan2022towards}, meta-learning techniques~\cite{fallah2020personalized,fallah2020convergence,jiang2019improving}, and other strategies. 
Our method can be considered as a form of the meta-learning approach. 

\subsection{Bayesian Federated Learning}
To address uncertainty estimation and overfitting with limited data, some studies have proposed Bayesian federated learning (BFL)~\cite{cao2023bayesian}. 
In BFL, incorporating suitable priors on model parameters, as regularization, mitigates overfitting with limited data. 
Additionally, the posterior equips the model with the capability to capture uncertainty. 
Consequently, BFL facilitates more robust and well-calibrated predictions~\cite{triastcyn2019federated,zhang2022personalized,liu2023bayesian,mou2023pfedv}. 
Recently, a cohort of BFL methods based on GPs has emerged~\cite{achituve2021personalized,dai2020federated,yin2020fedloc,yu2022federated}.
They utilize GP priors as the shared knowledge, leveraging the nonparametric nature of GPs to adapt more flexibly to complex data. 
However, the existing works seldom consider the coexistence of classification and regression tasks, a gap that this work seeks to address. 

\subsection{Multi-task Learning}
MTL~\cite{caruana1997multitask} has extensive applications across various domains, including natural language processing~\cite{collobert2008unified,dong2015multi}, computer vision~\cite{liu2019end,luo2012manifold}, recommendation systems~\cite{gao2023enhanced,li2020multi}, and more. 
Both MTL and FL involve knowledge transfer, but their focal points differ. 
MTL emphasizes leveraging correlations among multiple tasks~\cite{zhang2021survey,ruder2017overview}, 
while FL rigorously maintains client data privacy. 
Several works have adapted MTL methods to the FL domain while ensuring client data privacy~\cite{smith2017federated,marfoq2021federated,corinzia2019variational,dinh2021fedu,li2019online}. 
This work diverges from the existing works by employing a different emphasis. We adopt an MTL approach on clients, jointly modeling classification and regression tasks to facilitate knowledge transfer among different task types. 

\section{Preliminary}
\label{preliminary}
In this section, we show the basic concepts of GP regression and classification, MOGP, and P\'{o}lya-Gamma augmentation. 

\subsection{Gaussian Process Regression and Classification}
\label{gp_class}
GP regression is well-known for its flexibility and analytical inference. 
Specifically, the GP regression is formulated as: 
\begin{equation*}
    y(\mathbf{x})\mid f(\mathbf{x})\sim\mathcal{N}(f(\mathbf{x}),\sigma^2),
    f(\mathbf{x}) \sim \mathcal{GP}(m(\mathbf{x}),k(\mathbf{x},\mathbf{x}^\prime)),
\end{equation*}
where the output $y(\mathbf{x})$ is assumed to be obtained by an additive Gaussian noise, 
$\sigma^2$ is the noise variance treated as a hyperparameter; $m(\mathbf{x})$ is the GP mean function and $k(\mathbf{x},\mathbf{x}^\prime)$ is the GP kernel measuring data similarity. 
For GP regression, a notable advantage is analytical inference of posterior $f(\cdot)$ due to the Gaussian likelihood being conjugated to the GP prior. 
Moreover, if we aim to learn kernel hyperparameters from data, we can maximize the marginal likelihood which also possesses an analytical expression~\cite{rasmussen2003gaussian}. 

GP classification is more challenging. Here, we illustrate with the example of binary classification: 
\begin{equation*}
    y(\mathbf{x})\mid f(\mathbf{x})\sim \mathcal{B} (s(f(\mathbf{x}))),
    f(\mathbf{x}) \sim \mathcal{GP}(m(\mathbf{x}),k(\mathbf{x},\mathbf{x}^\prime)),
\end{equation*}
where $\mathcal{B}$ denotes the Bernoulli distribution (categorical distribution for multi-class classification), $s(\cdot)$ defines a link function: $\mathbb{R} \to (0,1)$ whose common choices include the cumulative distribution function of the standard Gaussian distribution (probit regression) and the sigmoid function (logistic regression). 
The primary challenge in GP classification lies in inference. Because the likelihood is non-conjugate to the GP prior, the posterior of the classification function $f(\cdot)$ lacks an analytical solution.
Normally, we resort to approximate inference such as MCMC, VI, and others. Additionally, the marginal likelihood is also intractable, making hyperparameter optimization difficult. 

\subsection{Multi-output Gaussian Processes}
MOGP~\cite{alvarezRL12kernel} extends GP to model multiple correlated output functions, providing a Bayesian nonparametric framework for multi-task learning. 
To define an MOGP, we need to establish a cross-covariance function representing the correlation among multiple outputs. Among various methods, we use the widely used linear model of coregionalization~\cite{journel1976mining}. 
Specifically, we assume each output function is a linear combination of $B$ basis functions drawn from $B$ independent GP priors: 
\begin{equation*}
    f_i(\mathbf{x})=\sum_{b=1}^B w_{i,b}g_b(\mathbf{x}),
    g_b(\mathbf{x})\sim\mathcal{GP}(m_b(\mathbf{x}),k_b(\mathbf{x},\mathbf{x}^\prime)),
\end{equation*}
where $f_i(\cdot)$ is the $i$-th output function, $g_b(\cdot)$ is the $b$-th basis function, $w_{i,b}\in\mathbb{R}$ is the mixing weight. As usual, $m_b(\cdot)$ is set to $0$, $k_b(\cdot,\cdot)$ is the kernel of the $b$-th GP. 
It is easy to see that the mean of $f_i(\cdot)$ is $0$, while the cross-covariance between two outputs is $k_{f_i,f_j}(\mathbf{x},\mathbf{x}')=\text{cov}[f_i(\mathbf{x}),f_j(\mathbf{x}')]=\sum_{b=1}^B w_{i,b}w_{j,b}k_b(\mathbf{x},\mathbf{x}')$. 
If we consider finite inputs, defining $\mathbf{f}_i$ as the function-value vector on the $i$-th task inputs, we obtain the discrete MOGP: $\mathbf{f}\sim\mathcal{N}(\mathbf{0},\mathbf{K})$, where $\mathbf{f}$ is the function-value vector of all tasks, 
$\mathbf{K}$ is a block matrix with each block denoted by $\mathbf{K}_{\mathbf{f}_{i},\mathbf{f}_{j}}$ where each entry is $k_{f_i,f_j}(\cdot,\cdot)$.

\subsection{P\'{o}lya-Gamma Augmentation}
Conducting effective posterior inference for GP classification has been a prominent focus within the Bayesian domain. 
Apart from directly employing MCMC or VI, several studies have proposed data augmentation methods that involve augmenting auxiliary latent variables into non-conjugate models, thereby transforming non-conjugate problems into conditional conjugate ones, 
and accelerating convergence compared to directly using MCMC or VI~\cite{wenzel2019efficient}. 
Here, we focus on the P\'{o}lya-Gamma augmentation for Bayesian logistic regression~\cite{polson2013bayesian}. 
The core of this method is the representation of the logistic likelihood as a mixture of Gaussians w.r.t. a P\'{o}lya-Gamma distribution. 
The definition of the P\'{o}lya-Gamma distribution is provided in~\cite{polson2013bayesian}, denoted as $p_{\text{PG}}(\omega\mid b,c)$, where $\omega\in \mathbb{R}^+$ with parameters $b>0$ and $c\in\mathbb{R}$. 
This work only requires its expectation $\mathbb{E}[\omega]=\frac{b}{2c}\tanh(\frac{c}{2})$. 

\section{Methodology}
\begin{figure*}[t]
    \centering
    \includegraphics[width=\textwidth]{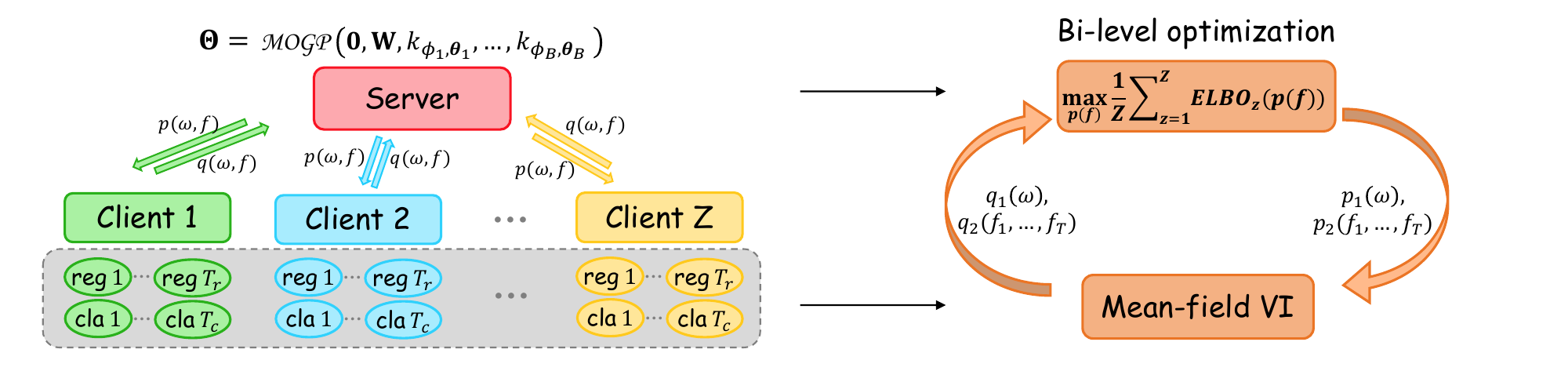}
    \caption{The overview of our model pFed-Mul. \textbf{Left}: System diagram. The central server aggregates the posteriors from local devices, updating a global MOGP prior redistributed for training local models. \textbf{Right}: Bi-level optimization. The subfigure illustrates an iterative application of mean-field VI at the local level and hyperparameter tuning at the global level.}
    \label{fig:framework}
    \Description{The overview of our model pFed-Mul. \textbf{Left}: System diagram. The central server aggregates the posteriors from local devices, updating a global MOGP prior redistributed for training local models. \textbf{Right}: Bi-level optimization. The subfigure illustrates an iterative application of mean-field VI at the local level and hyperparameter tuning at the global level.}
\end{figure*}

We delve into a personalized BFL model based on MOGP, with an overview outlined in \cref{fig:framework}.
In a distributed system comprising a single server and $\mathcal{Z}$ clients, where each client manages multiple correlated regression and classification tasks. For convenience, we assume an identical dataset size across all clients. 
On each client, we assume there are $T_r$ regression tasks with data $\mathcal{D}^r=\{{(\mathbf{x}_{i,n}^r,y_{i,n}^r)}_{n=1}^{N_i^r}\}_{i=1}^{T_r}$ and $T_c$ classification tasks with data $\mathcal{D}^c=\{{(\mathbf{x}_{i,n}^c,y_{i,n}^c)}_{n=1}^{N_i^c}\}_{i=1}^{T_c}$. $\mathbf{x}\in\mathcal{X}\subset\mathbb{R}^D$ represents the $D$-dim input. In regression, the output $y\in\mathbb{R}$, while in classification $y\in\{-1,1\}$\footnote{Here we focus on binary classification, while the extension to the multi-class case is discussed in \cref{multi_class}.}. 

\subsection{Client Level}
We present a MOGP-based multi-task learning model deployed on each client and detail optimization of the posterior distributions of latent functions.
\subsubsection{MOGP Model}
The correlation between classification and regression tasks is characterized by the MOGP prior and can be utilized to transfer knowledge, especially in scenarios with limited data~\cite{moreno2018heterogeneous}. Therefore, we obtain the Bayesian multi-task learning model based on MOGP on each client: 
\begin{subequations}
\label{eq1}
\begin{gather}
\label{eq1a}
\mathbf{y}^{r}\mid\{f_{i}^r\}_{i=1}^{T_r}\sim\prod_{i=1}^{T_r}\prod_{n=1}^{N_{i}^r}\mathcal{N}(f^r_{i,n},\sigma_i^2),\\
\label{eq1b}
\mathbf{y}^{c}\mid\{f_{i}^c\}_{i=1}^{T_c}\sim\prod_{i=1}^{T_c}\prod_{n=1}^{N^c_{i}}\mathcal{B}(s(y^c_{i,n}f^c_{i,n})),\\
\label{eq1c}
f_1,\ldots,f_T\sim \mathcal{MOGP}(0,\mathbf{W}, k_1,\ldots,k_B), 
\end{gather}
\end{subequations}
where \cref{eq1a} is the regression likelihood, \cref{eq1b} is the classification likelihood, and \cref{eq1c} is the MOGP prior; $f_i^r$ and $f_i^c$ refer to the respective $i$-th output function for the $T_r$ regression and $T_c$ classification tasks, 
$f_1,\ldots,f_T$ represent organizing all regression and classification functions together, thus $T=T_r+T_c$; 
$f^{\cdot}_{i,n}=f^{\cdot}_i(\mathbf{x}^{\cdot}_{i,n})$, $\mathbf{y}^{r}$ denotes all regression targets, $\mathbf{y}^{c}$ denotes all classification labels, $\mathbf{W}$ is the matrix of all mixing weights $w_{i,b}$, and $k_1,\ldots,k_B$ correspond to the kernels of $B$ basis functions. 
It is worth noting that we use logistic regression for classification tasks, meaning that the link function $s(\cdot)$ in \cref{eq1b} is sigmoid. This choice facilitates the use of P\'{o}lya-Gamma augmentation, simplifying the inference process afterward. 

\subsubsection{Posterior of Latent Functions}
Given the model provided in \cref{eq1}, the remaining task is to infer the posterior of each output function. 
For inference, as discussed in \cref{gp_class},  
the likelihood of classification tasks is not conjugate to the prior, resulting in non-analytical posteriors for $f_1,\ldots,f_T$. 
To address the non-conjugacy issue, many existing works employed Gaussian variational inference~\cite{hensman2015scalable,jahani2021multioutput}. This method assumes the variational distribution to be Gaussian, making the ELBO tractable. However, this method has drawbacks. On the one hand, it relies on parametric assumptions for the variational distribution, leading to increased approximation errors, especially when the true posterior deviates from Gaussian. 
On the other hand, due to the need to compute the expected log-likelihood in ELBO, which often requires Monte Carlo approximation, it typically exhibits low computational efficiency. 

To address the above issue, we adapt the P\'{o}lya-Gamma augmentation for MOGP to the federated setting. 
We augment the MOGP model with P\'{o}lya-Gamma random variables $\bm{\omega}$ for all classification tasks, one for each sample. Consequently, the original non-conjugate model $p(\mathbf{y}^{r},\mathbf{y}^{c},f_1,\ldots,f_T)$ is augmented to be a conditionally conjugate model $p(\mathbf{y}^{r},\mathbf{y}^{c},\bm{\omega},f_1,\ldots,f_T)$ allowing us to derive an analytical mean-field VI method. Following the common practice of mean-field VI, 
we approximate the true posterior in a factorized manner: $p(\bm{\omega},f_1,\ldots,f_T\mid\mathbf{y}^{r},\mathbf{y}^{c})\approx 
q(\bm{\omega},f_1,\ldots,f_T)=
q_1(\bm{\omega})q_2(f_1,\ldots,f_T)$. 
The optimal variational distribution is obtained by minimizing the Kullback-Leibler (KL) divergence between the factorized variational distribution and the true posterior, which is equivalent to the following optimization of ELBO: 
\begin{equation}
\begin{aligned}
    \max_{q(\bm{\omega},f)}
    \bigg\{&\mathbb{E}_{q(\bm{\omega},f)}[\log p(\mathbf{y}^{r}, \mathbf{y}^{c}\mid \bm{\omega}, \{f_{i}^r\}_{i=1}^{T_r}),\{f_{i}^c\}_{i=1}^{T_c})]\\
    &-\text{KL}(q(\bm{\omega}, f)\Vert p(\bm{\omega},f))\bigg\}. 
\end{aligned}
\label{equ:elbo_l}
\end{equation}
where $p(\omega,f)$ is the prior distribution distributed from server and fixed during local update. Specifically, the prior distribution is assumed as $p(\omega,f)=p(\omega)p(f)$ where $p(\omega)=p_\text{PG}(1,0)$ and $p(f)=\mathcal{MOGP}(0,\mathbf{W},k_1,\cdots,k_B)$. 
Under assumption of factorized variational distribution, we obtain the following local updates: 

\begin{subequations}
\label{equ:mf_local_q}
    \begin{equation}
    \label{equ:mf_local_q1}
        q_1(\bm{\omega}) = \prod_{i=1}^{T_c}\prod_{n=1}^{N_i^c}p_{\text{PG}}(\omega_{i,n}\mid 1, \tilde{f}_{i,n}^{c}), 
    \end{equation}
    \begin{equation}
    \label{equ:mf_local_q2}
        q_2(\mathbf{f}) = \mathcal{N}(\mathbf{m},\mathbf{\Sigma}), 
    \end{equation}
\end{subequations}
where $\tilde{f}_{i,n}^{c}=\sqrt{\mathbb{E}[f_{i,n}^{c^2}]}$, $\mathbf{\Sigma}=(\mathbf{H} + \mathbf{K}^{-1})^{-1}$, $\mathbf{m}=\mathbf{\Sigma}\mathbf{H}\mathbf{v}$,with $\mathbf{H}=\text{diag}(\mathbf{D}^r_{\cdot}, \mathbf{D}^c_{\cdot})$, $\mathbf{v}=[\mathbf{y}^r_{\cdot}, \frac{1}{2}{\mathbf{D}^{c}_{\cdot}}^{-1}\mathbf{y}^c_{\cdot}]^\top$, and $\mathbf{D}^r_i=\text{diag}(1/\sigma_i^2)$, $\mathbf{D}^c_i=\text{diag}(\mathbb{E}[\bm{\omega}_i])$. 
The detailed derivation of P\'{o}lya-Gamma augmentation and mean-field VI is provided in \cref{corollary1,proof:mf}. 

After obtaining the posterior distribution of $\mathbf{f}$, we can calculate the analytical expression for the predictive distribution at any point: 
\begin{equation}
\begin{aligned}
    q(f_i(x))&=\int p(f_i(x)\mid\mathbf{f}^i)q_2(\mathbf{f}^i)d\mathbf{f}^i=\mathcal{N}(\mu,\sigma^2), \\
    \mu&=\mathbf{k}_{\mathbf{x}^{\cdot}_i x}^{\top}\mathbf{K}_{\mathbf{x}^{\cdot}_i\mathbf{x}^{\cdot}_i}^{-1}\mathbf{m}_{\mathbf{x}_i^{\cdot}}, \\
    \sigma^2&=k_{xx}-\mathbf{k}_{\mathbf{x}_{i}^{\cdot}x}^{\top}\mathbf{K}_{\mathbf{x}^{\cdot}_i \mathbf{x}^{\cdot}_i}^{-1}\mathbf{k}_{\mathbf{x}^{\cdot}_i x}+\mathbf{k}_{\mathbf{x}^{\cdot}_i x}^{\top}\mathbf{K}_{\mathbf{x}^{\cdot}_i \mathbf{x}^{\cdot}_i}^{-1}\mathbf{\Sigma}_{\mathbf{x}^{\cdot}_i}\mathbf{K}_{\mathbf{x}^{\cdot}_i\mathbf{x}^{\cdot}_i}^{-1}\mathbf{k}_{\mathbf{x}_i x}.
\label{equ:predict}
\end{aligned}
\end{equation}

\subsection{Server Level}
The server maintains a global MOGP prior for the entire system, aggregates local posteriors to update the global MOGP prior, and distributes the updated global prior back to clients. 
The intuition behind our method is similar to that of pFedBayes~\citep{zhang2022personalized}. In practice, we often cannot directly assume a good prior suitable for the current data. 
As the communication rounds progress, the global MOGP becomes increasingly compatible with the data from all clients. 
This implies that we have found a relatively good prior. 
pFedBayes is a parametric method that assumes Gaussian variational distributions for each parameter, an assumption that may not always hold true. In contrast, our proposed method is non-parametric and imposes no assumptions on the form of the posterior distribution, with the only restriction being the independence between $f$ and $\bm{\omega}$. 

Specifically, at the server level, we aggregate the mean-field VI posteriors uploaded from clients and update the global MOGP prior by maximizing the averaged ELBO: 
\begin{equation}
\label{equ:agg}
\max_{p(f)}\frac{1}{\mathcal{Z}}\sum_{z=1}^{\mathcal{Z}}\text{ELBO}_z(p(f)), 
\end{equation}
where $\text{ELBO}_z$ represents the ELBO of the $z$-th client, which depends on the variational distribution $q$ and prior $p$. Since $q$ is uploaded by the client and fixed, the ELBO is solely a function of $p$. 
Thanks to the P\'{o}lya-Gamma augmentation, \cref{equ:agg} has an analytical solution, thus we can optimize the parameters of the prior, i.e., the kernel hyperparameters $\{\bm{\phi}_b\}_{b=1}^B$ pertain to B basis functions, the mixing weight $\mathbf{W}$, and the regression noise variance $\{\sigma_i^2\}_{i=1}^{T_r}$. The detailed derivation of \cref{equ:agg} is provided in \cref{analytic elbo}. 

For new incoming clients, based on the global MOGP served as a shared prior, the posterior of the classification and regression functions is further inferred with incorporation of their local data, which ensures personalization at the client level. 

\subsection{Deep Kernel and Inducing Points}
To further enhance the expressive capacity of MOGP, a deep kernel~\cite{wilson2016deep} is utilized in this study. The deep kernel involves a neural network $\eta(\cdot)$ with parameters $\bm{\theta}$ that transforms input data $x$ into a latent representation $\eta_{\bm{\theta}}(x)$. Subsequently, this representation is fed into a traditional kernel, thereby generating a new kernel: 
\begin{equation*}
k_{\bm{\phi},\bm{\theta}}(x_1,x_2) = \tilde{k}_{\bm{\phi}}(\eta_{\bm{\theta}}(x_1), \eta_{\bm{\theta}}(x_2)),
\label{deepkernel}
\end{equation*}
where $\tilde{k}_{\bm{\phi}}(\cdot,\cdot)$ is the base kernel, e.g., the radial basis function (RBF) kernel or others. 
One advantage of the deep kernel is its ability to learn a flexible input transformation metric in a data-driven manner, instead of relying directly on Euclidean distance based metrics that might not be suitable. 
For MOGP, $k_1,\ldots,k_B$ in \cref{eq1c} are modeled by deep kernels. Consequently, our model hyperparameters of prior $\bm{\Theta}$ include the kernel hyperparameters $\{\bm{\phi}_b,\bm{\theta}_b\}_{b=1}^B$, the mixing weight $\mathbf{W}$, and the regression noise variance $\{\sigma_i^2\}_{i=1}^{T_r}$. These hyperparameters are updated by maximizing averaged ELBO uploaded from each clients without alteration of the analytical solution in \cref{equ:agg}.

MOGP inherits GP's notorious cubic computational complexity w.r.t. the number of samples. 
The complexity of $O(N^3)$ becomes intolerable as the sample size $N$ per client increases. 
To enhance the computational efficiency, we employ the inducing points method~\cite{titsias2009variational}. 
We assume that these inducing inputs on each client are uniformly sampled from
local data and not uploaded to the server for aggregation, which upholds local privacy.
After introducing $M$ inducing points, 
the computational complexity decreases to $O(NM^2)$ ($M \ll N$), which is linear w.r.t. the number of samples on each client. 
The detailed derivation of mean-field VI with inducing points is provided in \cref{proof:mf_inducing}. 


\begin{algorithm}[!h]
\caption{pFed-Mul: Server}
    \begin{algorithmic}
        \STATE {\bfseries Input:} 
        server iteration $\mathcal{T}_s$, client size $\mathcal{Z}$, sample size $\mathcal{S}$ and initial global hyperparameters $\bm{\Theta}^{(g)}$, 
        
        \FOR{$t_s=0$ {\bfseries to} $\mathcal{T}_s-1$}
        \STATE $\mathbb{S}_{t_s}$ $\leftarrow$ Sample randomly the subset of clients with size $\mathcal{S}$
        \FOR{each client $z$ in $\mathbb{S}_{t_s}$}
        \STATE $\bm{\Theta}_z^{(l)}$ $\leftarrow$ Sent global hyperparameters $\bm{\Theta}^{(g)}$ to client $z$, 
        \STATE $q_{z,1}(\bm{\omega})$, $q_{z,2}(\mathbf{f})$ $\leftarrow$ Update local posteriors based on specific client data by \cref{alg:client}, 
        \ENDFOR
        \STATE $\bm{\Theta}^{(g)}$ $\leftarrow$ Optimize global MOGP prior according to \cref{equ:agg}. 
        \ENDFOR
    \end{algorithmic}
\label{alg:server}
\end{algorithm}
\begin{algorithm}[!h]
    \caption{pFed-Mul: Client}
    \begin{algorithmic}
        \STATE {\bfseries Input: }
        client iteration $\mathcal{T}_c$, 
        initial local hyperparameters $\bm{\Theta}_z^{(l)}$ for client $z$, 
        
        \FOR{$t_c = 0$ {\bfseries to} $\mathcal{T}_c-1$}
        \STATE $q_1(\bm{\omega})$ $\leftarrow$ Update variational distribution of P\'{o}lya-Gamma variables by \cref{equ:mf_local_q1},
        \STATE $q_2(\mathbf{f})$ $\leftarrow$ Update variational distribution of latent functions by \cref{equ:mf_local_q2},
        \ENDFOR
        \STATE $q(f(x)) \leftarrow$ Compute the predictive distribution of test points according to \cref{equ:predict}, 
    \end{algorithmic}
    \label{alg:client}
\end{algorithm}

\subsection{Algorithm}
In summary, at the client level, all clients receive the same global prior distributed by server, alternately update variational distributions $q(\bm{\omega})$ and $q(f)$ via \cref{equ:mf_local_q} to approximate posterior distributions based on the local data. 
At the server level, variational distributions $q(\bm{\omega})$ and $q(f)$ are aggregated and the averaged ELBO is optimized to update the glocal MOGP prior via \cref{equ:agg}. 
We term our method pFed-Mul whose pseudocode is provided in \cref{alg:server,alg:client}.

\section{Experiments}
\label{sec:experiment}
In this section, we utilize a synthetic dataset and two real-world datasets to showcase the performance of pFed-Mul in terms of accuracy, uncertainty estimation, and convergence. We did all experiments in this paper using servers with two GPUs (NVIDIA TITAN V with 12GB memory), two CPUs (each with 8 cores, Intel(R) Xeon(R) CPU E5-2620 v4 @ 2.10GHz), and 251GB memory.

\subsection{Experimental Setup}
\label{experiment_setup}
\subsubsection{Datasets}
We consider three datasets, including one synthetic dataset and two image datasets.

\textbf{Synthetic Data:} we assume that there exists 5 clients and each has one regression task and one classification task. The regression function $f_r$ and the classification function $f_c$ are assumed to be sampled from a MOGP on the domain $[0,100]$ with two kernels: $f_r,f_c\sim \mathcal{MOGP}(0,\mathbf{W},k_1,k_2)$. We select the RBF kernel $k(x_1, x_2) = \phi_0 \exp(-\frac{\phi_1}{2}\Vert x_1 - x_2\Vert_2^2)$. 
The regression function $f_r$ is used in \cref{eq1a} with a fixed noise variance $\sigma^2$ to sample the regression targets $\mathbf{y}_r$. 
The classification function $f_c$ is used in \cref{eq1b} to sample the classification labels $\mathbf{y}^{c}$. 
We simulate the synthetic data, where hyperparameters are $\sigma^2=0.1$, $\mathbf{W}=[[0.6, 0.4],[0.4, 0.6]]$, $\phi_0^{(1)}=1, \phi_0^{(2)}=2$, $\phi_1^{(1)}=0.02, \phi_1^{(2)}=0.01$. 

\textbf{CelebA:} 
this dataset comprises an extensive collection of over two million face images of celebrities, each accompanied by forty attribute annotations. The dataset exhibits a diverse range of images featuring significant variations in poses and background settings. Each image is associated with regression targets, such as the position of eyes, mouth, and classification labels such as the presence of eyeglasses, hair color, and smiling expressions. For more comprehensive details, readers are encouraged to refer to~\cite{liu2015faceattributes}. In our study, we specifically select the abscissa of the right side of the mouth as regression labels and whether or not the subject is smiling as classification labels. Given the close relationship between the position of the mouth corner and smiling, these two types of tasks have the potential to mutually transfer knowledge.

\textbf{Dogcat:} 
this dataset includes $20,000$ genuine images of dogs and cats and is widely employed for binary classification tasks in computer vision. The images in the dataset showcase various breeds of dogs and cats, captured in different poses, backgrounds, and lighting conditions. The primary objective of the dataset is to identify whether the images contain a dog or a cat, without the inclusion of regression labels. To create new regression labels, we introduce zero-mean Gaussian noise with a variance of $0.5$ into the original classification labels. As a result, regression labels exhibit bi-modal distribution. Specifically, for dog images, the regression targets are centered around $1$, while for cat images, they are centered around $-1$. It is evident that the classification labels and regression targets are closely related.
\begin{table*}[h]
  \centering
  \caption{The mean square error (MSE) for regression tasks and prediction accuracy (ACC) for classification tasks for all models. The experiments are conducted for two datasets, CelebA and Dogcat, under three few-shot scenarios, 10-shot 20-client, 20-shot 15-client, and 50-shot 10-client. FedPAC, pFedGP and pFedVEM are originally designed to process only the classification tasks, hence their results for regression tasks are not reported. The champion is highlighted in bold, runner-up with underline. \xmark indicates the model cannot handle this type of tasks.}
  \Description{The mean square error (MSE) for regression tasks and prediction accuracy (ACC) for classification tasks for all models. The experiments are conducted for two datasets, CelebA and Dogcat, under three few-shot scenarios, 10-shot 20-client, 20-shot 15-client, and 50-shot 10-client. FedPAC, pFedGP and pFedVEM are originally designed to process only the classification tasks, hence their results for regression tasks are not reported. The champion is highlighted in bold, runner-up with underline. \xmark indicates the model cannot handle this type of tasks.}
  \adjustbox{center}{
  \scalebox{0.95}{
  \begin{tabular}{ccccccc|cccccc}
  \toprule
    & \multicolumn{6}{c|}{\textbf{CelebA}} & \multicolumn{6}{c}{\textbf{Dogcat}} \\
    & \multicolumn{2}{c}{10-shot 20-client} & \multicolumn{2}{c}{20-shot 15-client} & \multicolumn{2}{c|}{50-shot 10-client} & \multicolumn{2}{c}{10-shot 20-client} & \multicolumn{2}{c}{20-shot 15-client} & \multicolumn{2}{c}{50-shot 10-client} \\
    &\textbf{MSE}($\downarrow$) & \textbf{ACC\%($\uparrow$)} & \textbf{MSE}($\downarrow$) & \textbf{ACC\%($\uparrow$)} & \textbf{MSE}($\downarrow$) & \textbf{ACC\%($\uparrow$)} & \textbf{MSE}($\downarrow$) & \textbf{ACC\%($\uparrow$)} & \textbf{MSE}($\downarrow$) & \textbf{ACC\%($\uparrow$)} & \textbf{MSE}($\downarrow$) & \textbf{ACC\%($\uparrow$)} \\
    \midrule
    FedAvg & 0.672 & 82.50 & 0.514 & 86.33 & 0.394 & 89.59 & \underline{0.667} & 94.70 & 0.576 & 94.63 & 0.515 & 97.13 \\
    FedPer & \textbf{0.369} & 79.04 & \underline{0.328} & 81.37 & \underline{0.261} & 86.68 & 0.731 & 95.40 & 0.682 & 96.92 & \underline{0.512} & 97.13 \\
    Scaffold & 0.774 & 77.36 & 0.649 & 79.32 & 0.545 & 85.12 & 0.720 & 94.41 & 0.667 & 96.77 & 0.541 & 97.43 \\
    pFedMe & 0.792 & 78.04 & 0.657 & 79.84 & 0.552 & 85.44 & 0.751 & 94.60 & 0.673 & 96.82 & 0.543 & 97.13 \\
    FedPAC & \xmark & 77.81 & \xmark & 79.17 & \xmark & 81.60 & \xmark & 96.72 & \xmark & \textbf{97.51} & \xmark & 97.96 \\
    \midrule
    pfedGP & \xmark & 76.96 & \xmark & 87.95 & \xmark & 89.92 & \xmark & 92.67 & \xmark & 97.41 & \xmark & \underline{98.17} \\
    pFedVEM & \xmark & 78.91 & \xmark & 80.47 & \xmark & 84.12 & \xmark & 95.03 & \xmark & 95.55 & \xmark & 97.32 \\
    \midrule
    pFed-St & 0.690 & \underline{83.80} & \textbf{0.321} & \underline{88.31} & \textbf{0.221} & \underline{90.28} & 0.799 & \underline{96.83} & \underline{0.570} & 96.92 & 0.525 & 97.82 \\
    \textbf{pFed-Mul} & \underline{0.488} & \textbf{86.36} & 0.476 & \textbf{88.47} & 0.301 & \textbf{90.76} & \textbf{0.512} & \textbf{96.88} & \textbf{0.422} & \underline{97.46} & \textbf{0.398} & \textbf{98.22} \\
    \bottomrule
    \end{tabular}}}
  \label{tab:metric}
\end{table*}

\subsubsection{Baselines}
We compare our pFed-Mul with competitive FL methods, which can be categorized into two groups: (1) Bayesian FL methods, \textbf{pFedGP}~\cite{achituve2021personalized} and \textbf{pFedVEM}~\cite{zhu2023confidence}; (2) frequentist FL methods, \textbf{FedAvg}~\cite{mcmahan2017communication}, \textbf{FedPer}~\cite{arivazhagan2019federated}, \textbf{Scaffold}~\cite{karimireddy2020scaffold}, \textbf{pFedMe}~\cite{t2020personalized} and \textbf{FedPAC}~\cite{xu2023personalized}. 
As the existing methods are designed for single task, we implement them separately for each type of task and present the respective outcomes. 
Moreover, we introduce an additional single-task version of pFed-Mul, denoted as \textbf{pFed-St}, which is exclusively designed to handle a single type of tasks.

\subsubsection{Training Protocol}
For the synthetic dataset, at the server level, we assume a global MOGP prior with two RBF kernels without deep architecture and distribute it to each client. 
At the client level, posterior distributions are updated via mean-field VI and sent back to the server for optimizing the averaged ELBO w.r.t. hyperparameters $\mathbf{W}$, $\bm{\phi}$, and $\sigma^2$. 
We initialize all hyperparameters as the ground truth.
The number of global communication rounds, mean-field iterations and local updates are set to 20, 2 and 2, respectively.

Similarly, for the real-world datasets, we assume that each client has one regression task and one classification task. The training data are partitioned in a non-overlapping manner and distributed to individual clients. 
It is worth noting that this setup is designed for computational convenience, but our method can adapt to scenarios involving multiple tasks (more than two) per client and task heterogeneity among clients. 
A MOGP prior with two deep kernels is employed where RBF serves as the base kernel. 
The deep architecture $\eta_{\bm{\theta}}(\cdot)$ in the deep kernel is implemented using ResNet-18~\cite{he2016deep}.
The initial hyperparameters are set as follows, $\phi_0^{(1)}=\phi_0^{(2)}=1,\phi_1^{(1)}=\phi_1^{(2)}=0.01,\sigma^2=0.1$, and $\mathbf{W}$ is tuned with fixed other hyperparameters.
The number of global communication rounds, mean-field iterations and local updates are set to 70, 2 and 2, respectively. To demonstrate the advantage of our model, all real-world data experiments are conducted in few-shot settings where each client possesses only limited data.


Furthermore, we have the option to update global MOGP prior by optimizing summation of ELBOs from a selection of clients according to \cref{equ:agg}. Alternatively, we can update certain hyperparameters by \cref{equ:agg}, while retaining others that are optimized by client specific ELBOs.
This strategy is designed to improve the level of personalization for the clients. Specifically, we update all hyperparamters $\mathbf{W}$, $\bm{\phi}$, $\sigma^2$ of global prior for synthetic dataset via \cref{equ:agg}, while solely backbone $\bm{\theta}$ for real image datasets with others optimized locally.

\begin{figure}[t]
\centering
    \subcaptionbox{pFed-Mul (left:regression right:classification)\label{fig:syn-mul}}{\includegraphics[width=0.23\textwidth]{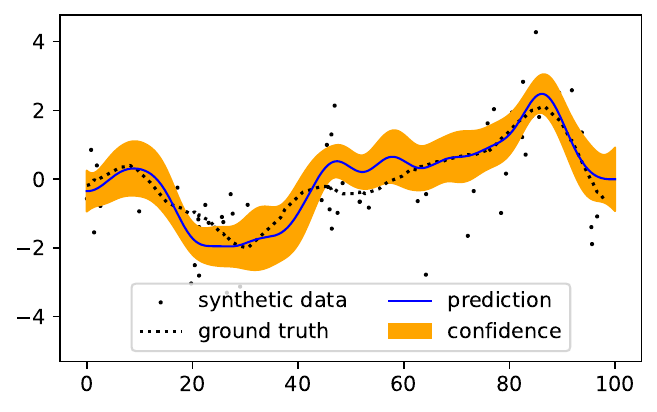}
    \includegraphics[width=0.23\textwidth]{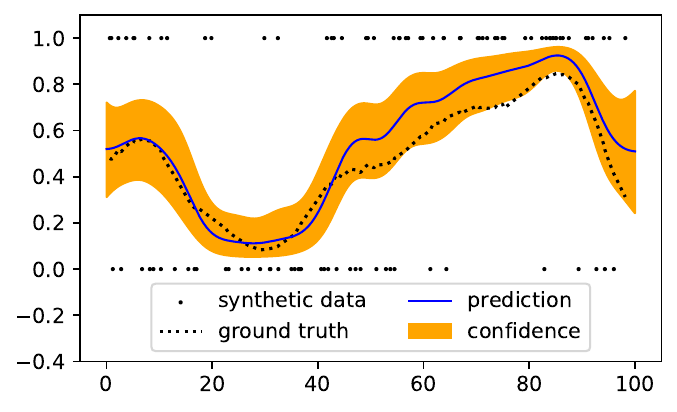}}
    \subcaptionbox{pFed-St (left:regression right:classification)\label{fig:syn-st}}{\includegraphics[width=0.23\textwidth]{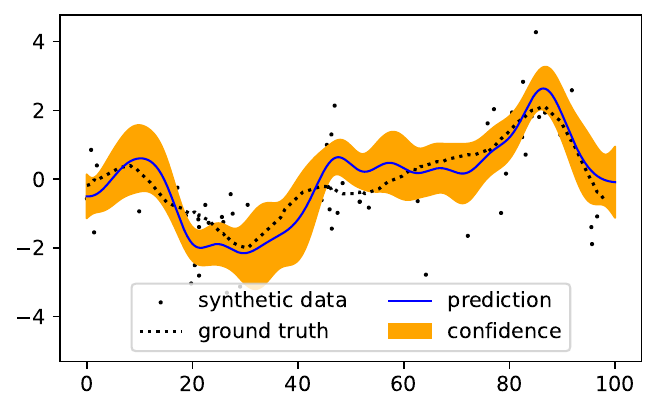}
    \includegraphics[width=0.23\textwidth]{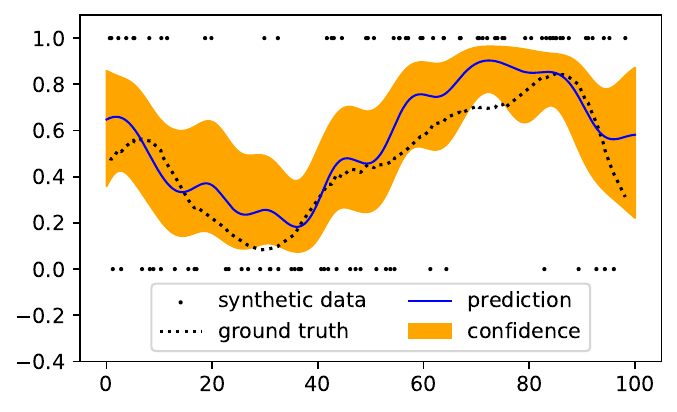}}
    \caption{The estimated posterior of latent functions from pFed-Mul and pFed-St on one client. 
    pFed-Mul, achieves a better fit, especially for classification. 
    Compared with pFed-St, pFed-Mul enables the transfer of knowledge from other task types, effectively reducing uncertainty, i.e. posterior variance (orange areas).}
    \label{fig:synthetic}
    \Description{The estimated posterior of latent functions from pFed-Mul and pFed-St on one client. 
    pFed-Mul, achieves a better fit, especially for classification. 
    Compared with pFed-St, pFed-Mul enables the transfer of knowledge from other task types, effectively reducing uncertainty, i.e. posterior variance (orange areas).}
\end{figure}

\begin{figure*}[h]
\centering
\begin{subfigure}[b]{0.24\textwidth}
    \centering
    \includegraphics[width=\textwidth]{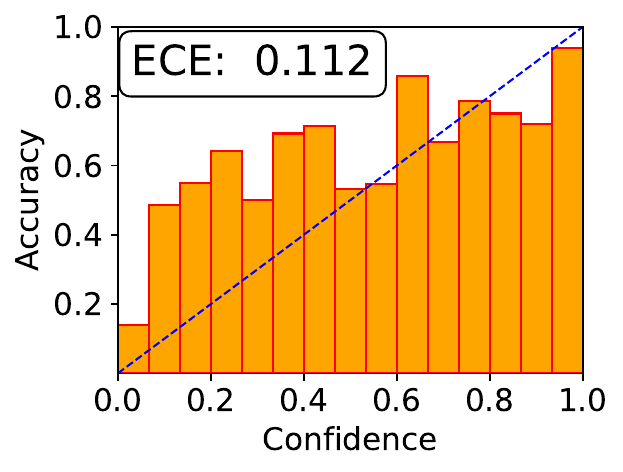}
    \caption{FedAvg}
    \label{fig:cali_fedavg}
\end{subfigure}
\begin{subfigure}[b]{0.24\textwidth}
    \centering
    \includegraphics[width=\textwidth]{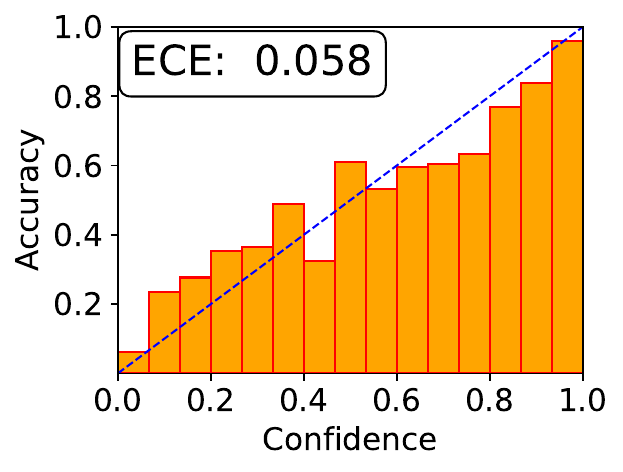}
    \caption{FedPer}
    \label{fig:cali_fedper}
\end{subfigure}
\begin{subfigure}[b]{0.24\textwidth}
    \centering
    \includegraphics[width=\textwidth]{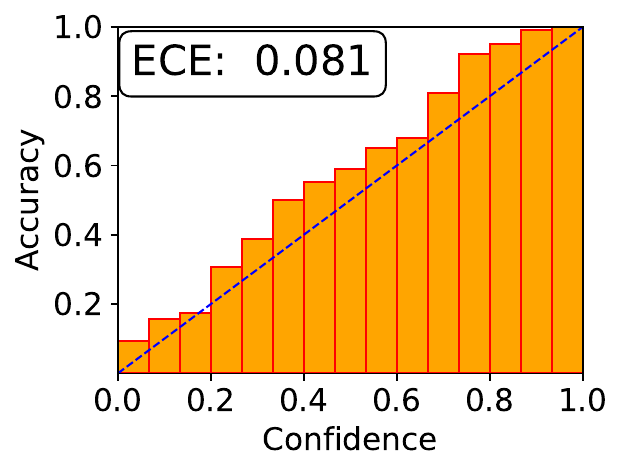}
    \caption{Scaffold}
    \label{fig:cali_scaffold}
\end{subfigure}
\begin{subfigure}[b]{0.24\textwidth}
    \centering
    \includegraphics[width=\textwidth]{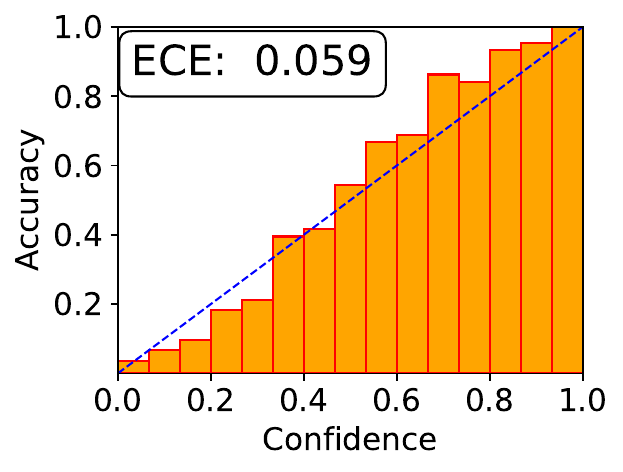}
    \caption{pFedMe}
    \label{fig:cali_pFedMe}
\end{subfigure}
\\
\begin{subfigure}[b]{0.24\textwidth}
    \centering
    \includegraphics[width=\textwidth]{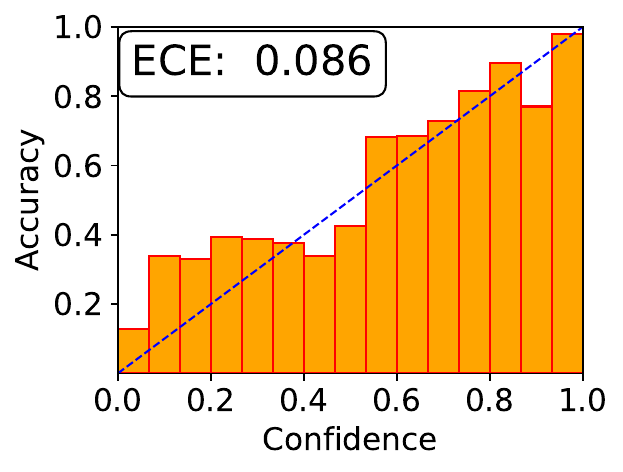}
    \caption{FedPAC}
    \label{fig:cali_FedPAC}
\end{subfigure}
\begin{subfigure}[b]{0.24\textwidth}
    \centering
    \includegraphics[width=\textwidth]{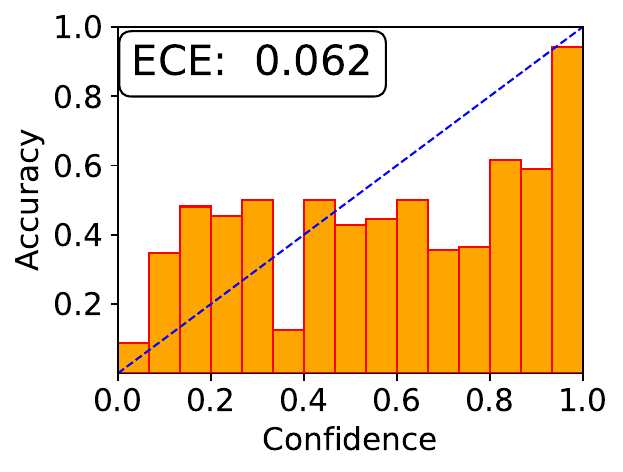}
    \caption{pFedGP}
    \label{fig:cali_pFedGP}
\end{subfigure}
\begin{subfigure}[b]{0.24\textwidth}
    \centering
    \includegraphics[width=\textwidth]{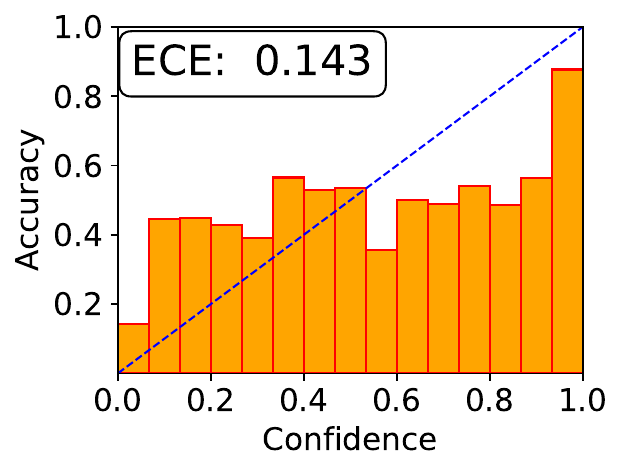}
    \caption{pFedVEM}
    \label{fig:cali_pFedVEM}
\end{subfigure}
\begin{subfigure}[b]{0.24\textwidth}
    \centering
    \includegraphics[width=\textwidth]{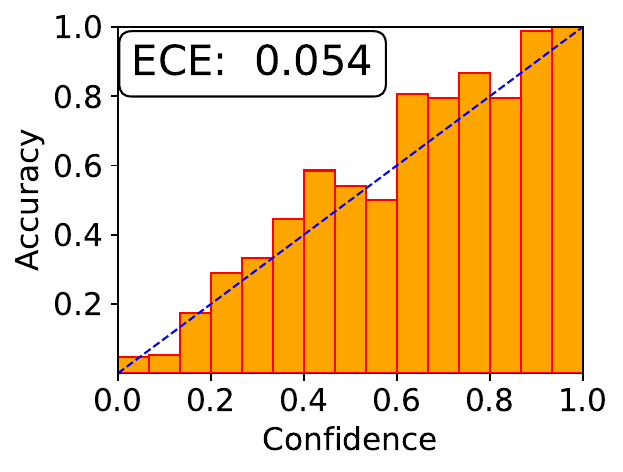}
    \caption{pFed-Mul}
    \label{fig:cali_pFed-Mul}
\end{subfigure}
\caption{Reliability diagrams for all methods. We plot the perfect calibration as blue diagonals, and practical result as orange bars. The disparity between the top of orange bars and blue line represents the degree of calibration, with the expected calibration error (ECE) calculated for comparison and placed in the top-left corner of diagrams. Our proposed method, pFed-Mul, demonstrates best calibration performance, ranking first in terms of ECE.}
\label{fig:calibration}
\Description{Reliability diagrams for all methods. We plot the perfect calibration as blue diagonals, and practical result as orange bars. The disparity between the top of orange bars and blue line represents the degree of calibration, with the expected calibration error (ECE) calculated for comparison and placed in the top-left corner of diagrams. Our proposed method, pFed-Mul, demonstrates best calibration performance, ranking first in terms of ECE.}
\end{figure*}

\subsection{Performance of Prediction}
\label{prediction}

\subsubsection{Synthetic Data} 

We conduct a visual analysis to compare the estimated posterior of latent functions from pFed-Mul with that from pFed-St on one client in \cref{fig:synthetic}. 

As shown in \cref{fig:syn-mul}, the results indicate that our proposed method successfully recovers the ground-truth latent functions. 
Furthermore, by comparing pFed-Mul to pFed-St (\cref{fig:syn-st}) that handle only one type of tasks, we summarize key findings as follows.
(1) We observe that pFed-Mul improves the fitting of latent functions, especially for the classification functions. 
The more significant improvement for the classification functions can be attributed to the fact that the target values of regression functions exhibit greater volatility, making them relatively easier to estimate. Conversely, the target values of classification functions, passed through a sigmoid function, are compressed within the range of $[0,1]$, thereby making their estimation more challenging. 
(2) For pFed-St, a smaller data size results in greater uncertainty in parameter estimation (posterior variance), while pFed-Mul facilitates knowledge transfer across different task types, thereby reducing such uncertainty. 
These outcomes show the necessity of knowledge transfer among diverse task types, particularly in few-shot scenarios.

\subsubsection{Real Data}  
We conduct experiments on CelebA and Dogcat, in three different settings: 10-shot individually among 20 clients, 20-shot individually among 15 clients, and 50-shot individually among 10 clients. 
The evaluation metrics including mean square error (MSE) for regression tasks and prediction accuracy (ACC) for classification tasks are computed for all methods. 

The results, summarized in \cref{tab:metric}, show that, (1) pFed-Mul consistently outperforms existing methods across almost all scenarios. This observation showcases remarkable adaptability of our proposed method from synthetic datasets to intricate real datasets. 
In terms of evaluation metrics, the most significant improvements observed in regression and classification tasks amount to $0.155$ and $3.86\%$ respectively. 
(2) In comparison to the single-task baseline models, the utilization of the multi-task framework demonstrates an increase of accuracy in both regression and classification tasks, highlighting the advantage of multi-task learning, particularly with limited data. 
This success can be attributed to two aspects. Firstly, incorporating more tasks enables the utilization of additional data, mitigating local overfitting and enhancing global robustness. Secondly, leveraging prior knowledge among tasks achieves better prior distribution and enhances convergence efficiency.

\subsection{Performance of Uncertainty Estimation}
We illustrate that our method can qualify uncertainty and achieve superior performance to previous baselines in terms of model calibration and OOD detection. These evaluations are conducted in a setting of $50$-shot individually among $10$ clients. 

\subsubsection{Model Calibration}

We assess uncertainty by calibrating the binary classification tasks for CelebA. 
The reliability diagrams, as depicted in \cref{fig:calibration}, showcase the disparity between the perfect calibration (blue diagonals) and the model's calibration (orange bars). 
To quantitatively compare the calibration, we calculate the expected calibration error (ECE), which measures weighted average between empirical accuracy and model's confidence as suggested in~\cite{guo2017calibration}.
The results indicate that pFed-Mul demonstrates calibration performance superior to the baseline models. Specifically, pFed-Mul ranks first in terms of ECE, FedPer exhibits runner-up performance, and pFedVEM performs worst among all baselines.

\subsubsection{OOD Detection}

\begin{figure}[t]
    \centering
    \begin{subfigure}[b]{0.48\textwidth}
        \centering
        \includegraphics[width=\textwidth]{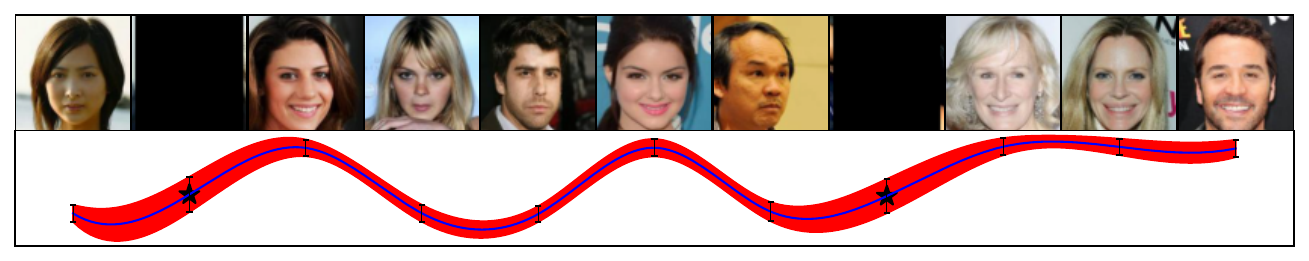}
        \label{fig:ood_celeba}
        \caption{CelebA}
    \end{subfigure}
    \begin{subfigure}[b]{0.48\textwidth}
        \centering
        \includegraphics[width=\textwidth]{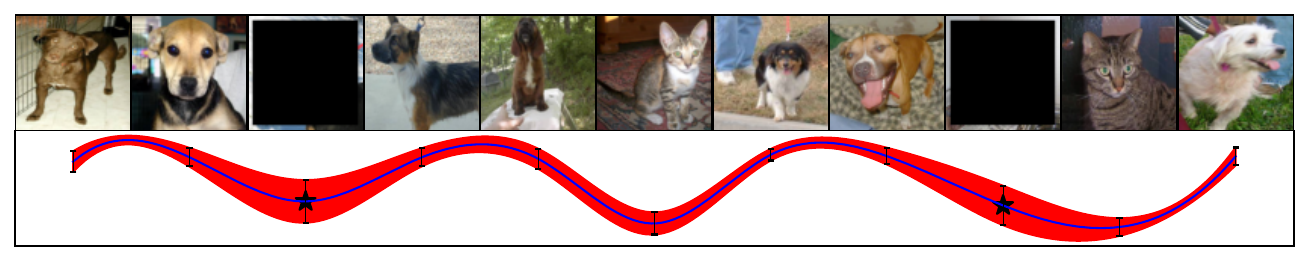}
        \label{fig:ood_dogcat}
        \caption{Dogcat}
    \end{subfigure}
    \caption{OOD detection for CelebA and Dogcat. The predictive mean and variance of latent functions are depicted by blue lines and red areas beneath each image respectively. Positions where the image is masked as an OOD sample are denoted by black stars. A greater variance (wider area) is observed for OOD samples.}
    \Description{OOD detection for CelebA and Dogcat. The predictive mean and variance of latent functions are depicted by blue lines and red areas beneath each image respectively. Positions where the image is masked as an OOD sample are denoted by black stars. A greater variance (wider area) is observed for OOD samples.}
    \label{fig:ood}
\end{figure}

The uncertainty of prediction provided by the Bayesian framework is crucial for detecting OOD samples. To demonstrate this, we select a series of samples from CelebA and Dogcat, randomly mask two of them, and compute the predictive variance in classification tasks. The results are depicted in \cref{fig:ood}. 
It is evident that the masked images demonstrate a larger semantic shift compared to in-distribution images. Therefore, we observe a greater predictive variance (depicted as red areas) under them. 
This visualization highlights the robustness of our method: pFed-Mul not only provides predictions but also outputs the uncertainty of predictions. When the uncertainty is large, it indicates that the model is not confident in the predicted results. 


\subsection{Convergence Rate}

\begin{figure}[t]
\centering
    \subcaptionbox{CelebA\label{fig:effi_celeba}}{\includegraphics[width=0.23\textwidth]{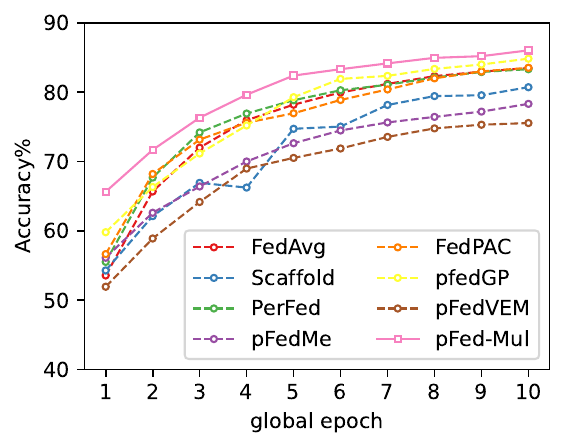}}
    \subcaptionbox{Dogcat\label{fig:effi_dogcat}}{\includegraphics[width=0.23\textwidth]{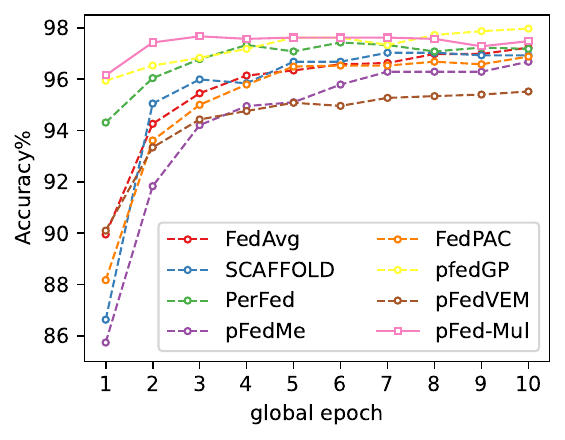}}
    \caption{Convergence rate of all models on both datasets. pFed-Mul consistently converges to a comparable test accuracy plateau with a remarkable convergence rate.}
\Description{Convergence rate of all models on both datasets. pFed-Mul consistently converges to a comparable test accuracy plateau with a remarkable convergence rate.}
\label{fig:convergence}
\end{figure}

We conduct a comparison between pFed-Mul and other baselines about convergence rate. 
For all models, in each communication round, we assume that the local parameters/variational distributions are updated $2$ times before being uploaded to the server in a setting of 50-shot individually among 10 clients. The convergence curve of test accuracy for classification tasks within the initial $10$ communication rounds is depicted in \cref{fig:convergence}. 

In \cref{fig:effi_celeba}, pFed-Mul consistently converges to the best test accuracy plateau after 10 global rounds of communication, with a remarkable convergence rate. Meanwhile, in \cref{fig:effi_dogcat}, pFed-Mul not only outperforms other methods in the initial rounds, showing a substantial lead over the runner-up, pFedGP, but also maintains stable performance comparable to other approaches. 
The superior convergence rate of pFed-Mul stems from our adoption of P\'{o}lya-Gamma augmentation for classification tasks. As proven in \cite{hoffman2013stochastic}, employing mean-field VI for a conditionally conjugate model is equivalent to optimizing the ELBO using natural gradient descent~\citep{amari1998natural} with step size of $1$. This second-order optimization method exhibits an improved convergence rate compared to traditional first-order optimization methods.

Beyond the convergence rate, the test accuracy convergence curve of pFed-Mul exhibits a stable monotonic increase without significant fluctuations, indicating remarkable stability.
Both convergence rate and stability, hold paramount importance for a model's adaptability in real-world scenarios, emphasizing training efficiency, low latency, and remarkable performance. 

\subsection{Ablation Studies}
We conduct ablation studies to assess various model components in the setting of 50-shot individually among 10 clients, enhancing our comprehension of the model's behavior. 

\paragraph{Aggregated Hyperparameters}
In the implementation, we can optimize only specific hyperparameters by \cref{equ:agg}, leaving the rest optimized by local ELBOs, thereby enhancing personalization for the clients. 
To investigate this, we compare several versions of pFed-Mul: pFed-Mul-N which optimizes the parameters of the neural network in the deep kernel $\bm{\theta}$ on server (the one we use in \cref{prediction}); 
pFed-Mul-K which optimizes all kernel hyperparameters $\bm{\phi},\bm{\theta}$ on server; pFed-Mul-W which optimizes all kernel hyperparameters and mixing weights $\bm{\phi},\bm{\theta},\mathbf{W}$ on server; pFed-Mul-A which optimizes all hyperparameters $\bm{\phi},\bm{\theta},\mathbf{W},\sigma^2$ on server. 
The results are shown in \cref{tab:ablation}. We can see that pFed-Mul-N strikes a balance between local personalization and global generalization, outperforming other versions. 
pFed-Mul-A performs unsatisfying, underscoring the necessity of personalization in FL. 

\begin{table}
  \centering
  \caption{The prediction performance for ablation studies. In the first block, we analyze different levels of personalization with various optimized hyperparameters. In the second block, we conduct experiments with different base kernels. In the third block, we compare different backbones.}
  \Description{The prediction performance for ablation studies. In the first block, we analyze different levels of personalization with various optimized hyperparameters. In the second block, we conduct experiments with different base kernels. In the third block, we compare different backbones.}
    \scalebox{0.9}{\begin{tabular}{ccccc}
    \toprule
    & \multicolumn{2}{c}{\textbf{CelebA}} & \multicolumn{2}{c}{\textbf{Dogcat}} \\
    &\textbf{MSE}($\downarrow$) & \textbf{ACC\%($\uparrow$)} & \textbf{MSE}($\downarrow$) & \textbf{ACC\%($\uparrow$)}\\
    \midrule
    & \multicolumn{4}{c}{\textbf{Aggregated Hyperparameters}} \\
    \midrule
    pFed-Mul-K & \underline{0.313} & \underline{88.56} & 0.659 & \underline{98.12} \\
    pFed-Mul-W & 0.449 & 88.04 & 0.426 & 97.82 \\
    pFed-Mul-A & 0.466 & 87.52 & \underline{0.417} & 97.92 \\
    \textbf{pFed-Mul-N} & \textbf{0.301} & \textbf{90.76} & \textbf{0.398} & \textbf{98.22} \\
    \midrule
    & \multicolumn{4}{c}{\textbf{Base Kernel}} \\
    \midrule
    Linear Kernel & 0.476 & 85.80  & \underline{0.442} & 96.98 \\
    Laplace Kernel & 0.436 & 90.36 & 0.485 & \underline{97.87} \\
    Cauchy Kernel & \underline{0.385} & \underline{90.40}  & 0.453 & \underline{97.87} \\
    \textbf{RBF Kernel} & \textbf{0.301} & \textbf{90.76} & \textbf{0.398} & \textbf{98.22} \\
    \midrule
    & \multicolumn{4}{c}{\textbf{Backbone}} \\
    \midrule
    EfficientNet & \underline{0.306} & \underline{89.64} & \underline{0.405} & \underline{97.67} \\
    ShuffleNet & 0.396 & 87.44 & 0.421 & 96.09 \\
    RegNet & \textbf{0.301} & 89.00 & 0.409 & \textbf{98.22}\\
    \textbf{ResNet} & \textbf{0.301} & \textbf{90.76} & \textbf{0.398} & \textbf{98.22} \\
    \bottomrule
    \end{tabular}}
  \label{tab:ablation}%
\end{table}

\paragraph{Base Kernel}
The base kernels in MOGP also have a significant impact on the results. 
We compare the MOGP models with linear kernel, Laplace kernel, Cauchy kernel, and RBF kernel. 
The expressions for all kernels are shown in \cref{app.experiment}. 
The results are shown in \cref{tab:ablation}, and reveal that the RBF kernel stands out as the best-performing kernel, consistent with previous studies. Additionally, the Cauchy kernel achieves a runner-up position, demonstrating results comparable to the RBF kernel. 
In contrast, the linear kernel exhibits inferior performance.

\paragraph{Backbone}
Recalling that we employ ResNet-18 as the backbone in deep kernels, it is necessary to analyse the impact of backbone on the prediction performance. Therefore, we replace ResNet-18 with EfficientNet-B2~\cite{tan2019efficientnet}, ShuffleNet-v2-2x~\cite{ma2018shufflenet}, RegNet-Y-1.6GF~\cite{radosavovic2020designing} and report results on both dataset in \cref{tab:ablation}, where the amounts of parameters of all backbones are comparable. 
The results demonstrate that it is beneficial for prediction to utilize ResNet-18 as the feature extractor. Meanwhile, ShuffleNet exhibits worst performance among all backbones.

\section{Conclusion}
In summary, our approach addresses a crucial limitation in FL, considering task diversity on clients. 
The proposed approach integrates multi-task learning using MOGP at the local level and federated learning at the global level. 
MOGP is effective in handling correlated classification and regression tasks, providing a Bayesian non-parametric framework that inherently quantifies uncertainty. 
To overcome challenges in posterior inference, we employ the P\'{o}lya-Gamma augmentation technique, leading to an analytical mean-field VI. 
The experimental results demonstrate our method's superiority in predictive performance, uncertainty calibration, OOD detection and convergence rate. 
The results highlight the method's potential across diverse applications.

\section*{Acknowledgments}
This work was supported by NSFC Projects (Nos. 62106121, 72171229), the MOE Project of Key Research Institute of Humanities and Social Sciences (22JJD110001), the Big Data and Responsible Artificial Intelligence for National Governance, Renmin University of China, the fundamental research funds for the central universities, and the research funds of Renmin University of China (24XNKJ13).

\bibliographystyle{ACM-Reference-Format}
\balance
\bibliography{sample-base}

\newpage
\appendix

\section{Classification Likelihood with P\'{o}lya-Gamma Augmentation}
\label{corollary1}
\begin{proof}
In accordance with Theorem 1 in \cite{polson2013bayesian}, the likelihood of classification task is delineated as follows, 
\begin{equation*}
p(\mathbf{y}_i^c\mid \mathbf{f}_i^c)=\prod_{n=1}^{N_i^c}\int e^{h(\omega_{i,n}, y_{i,n}^c, f_{i,n}^c)}p_{\text{PG}}(\omega_{i,n}\mid 1,0)d\omega_{i,n}, 
\end{equation*}
where $h(\omega_{i,n}, y_{i,n}^c, f_{i,n}^c)=\frac{1}{2}y_{i,n}^cf_{i,n}^c-\frac{1}{2}\omega_{i,n}{f_{i,n}^c}^2-\log2$. Hence, the augmented likelihood is, 
\begin{equation}
\begin{aligned}
    p(\mathbf{y}_i^c, \bm{\omega}_i\mid\mathbf{f}_i^c)&=\prod_{n=1}^{N_i^c}e^{\frac{1}{2}y_{i,n}^cf_{i,n}^c-\frac{1}{2}\omega_{i,n}{f_{i,n}^c}^2-\log 2}p_{\text{PG}}(\omega_{i,n}\mid 1,0)\\
    &\propto e^{\frac{1}{2}\mathbf{y}_i^{c\top}\mathbf{f}_i^c-\frac{1}{2}\mathbf{f}_i^{c\top}\text{diag}(\bm{\omega}_i)\mathbf{f}_i^{c}}\prod_{n=1}^{N_i^c}p_{\text{PG}}(\omega_{i,n}\mid 1,0). 
\end{aligned}
\end{equation}
Of particular note is the exponential term within the final equation, which is demonstrably proportional to the Gaussian distribution $\mathcal{N}(\mathbf{f}_i^c\mid\frac{1}{2}\text{diag}(\bm{\omega}_i)^{-1}\mathbf{y}_i^c, \text{diag}(\bm{\omega}_i)^{-1})$. Therefore, augmented likelihood is conditionally conjugate to the MOGP prior.     
\end{proof}

\section{Proof of Mean-field VI without Inducing Points}
\label{proof:mf}
\begin{proof}
Consider first the factorized condition of variational distributions in mean-field VI, $q(\bm{\omega},f)=q(\bm{\omega})q(f)$, hence, \cref{equ:mf_local_q} is rewritten as, 
\begin{equation}
\label{equ:elbo}
\begin{aligned}
    \max_{q(\bm{\omega}),q(f)}\bigg\{&\mathbb{E}_{q(f_{\cdot}^{r})}[\log p(\mathbf{y}^{r} \mid f_{\cdot}^r)]+\mathbb{E}_{q(\bm{\omega})q(f_{\cdot}^c)}[\log p(\mathbf{y}^{c}\mid \bm{\omega},f_{\cdot}^c)]\\
    &-\text{KL}(q(f)\Vert p(f))-\text{KL}(q(\bm{\omega})\Vert p(\bm{\omega}))\bigg\},
\end{aligned}
\end{equation}
where $p(f)=\mathcal{MOGP}(\bm{0}, \mathbf{W},k_1,\cdots,k_B)$ with discrete version $p(\mathbf{f})=\mathcal{N}(0,\mathbf{K})$ on all observed samples, and $p(\omega_{i,n})=p_\text{PG}(1,0)$.  
The optimal variational distribution is subsequently obtained: 
\begin{subequations}
\begin{equation}
    q_1(\bm{\omega})\propto e^{\mathbb{E}_{q_2(\mathbf{f})}\log p(\mathbf{y}^r, \mathbf{y}^c, \bm{\omega}, \mathbf{f})}, 
    \label{mf-omega}
\end{equation}
\begin{equation}
    q_2(\mathbf{f})\propto e^{\mathbb{E}_{q_1(\bm{\omega})}\log p(\mathbf{y}^r, \mathbf{y}^c, \bm{\omega}, \mathbf{f})}. 
    \label{mf-f}
\end{equation}
\end{subequations}

Besides, we can write detailed expression of the joint distribution by applying augmented classification likelihood in \cref{corollary1}. 
\begin{equation}
\begin{aligned}    &p(\mathbf{y}^{r},\mathbf{y}^{c},\bm{\omega},\mathbf{f})
=p(\mathbf{y}^{r}\mid\{\mathbf{f}_{i}^r\}_{i=1}^{T_r})p(\mathbf{y}^c,\bm{\omega}\mid\{\mathbf{f}^c_i\}_{i=1}^{T_c})p(\mathbf{f})\\
=&\prod_{i=1}^{T_r}\prod_{n=1}^{N_i^r}\mathcal{N}(y_{i,n}^r\mid f_{i,n}^r, \sigma^2_i)\cdot\prod_{i=1}^{T_c}\prod_{n=1}^{N_i^c}e^{\frac{1}{2}y_{i,n}^cf_{i,n}^c-\frac{1}{2}\omega_{i,n}{f_{i,n}^c}^2-\log 2}\\
&\cdot p_{\text{PG}}(\omega_{i,n}\mid 1,0)\cdot p(\mathbf{f}).
\end{aligned}
\label{joint-dist}
\end{equation}
Substituting \cref{joint-dist} into \cref{mf-omega,mf-f}, closed-form solutions for both are derived respectively in the following. 

\textbf{Derivation for \cref{mf-omega}}: Substituting \cref{joint-dist} into \cref{mf-omega} and remaining terms that contain factors $\bm{\omega}$, the optimal distribution is derived as: 
\begin{equation}
\label{equ:mf-omega}
\begin{aligned}
    q_1(\bm{\omega})&\propto\prod_{i=1}^{T_c}\prod_{n=1}^{N_i^c}e^{-\frac{1}{2}\omega_{i,n}\mathbb{E}[{f_{i,n}^c}^2]}p_{\text{PG}}(\omega_{i,n}\mid 1,0)\\
    &\propto\prod_{i=1}^{T_c}\prod_{n=1}^{N_i^c}p_{\text{PG}}(\omega_{i,n}\mid 1, \tilde{f}_{i,n}^c),  
\end{aligned}
\end{equation}
the last line is derived by 
$p_{\text{PG}}(\omega\mid b,c) = \frac{\exp(-\frac{c^2}{2}\omega)p_{\text{PG}}(\omega\mid b, 0)}{E_{\omega\sim p_{\text{PG}}(\omega \mid b,0)}\{\exp (-\frac{c^2}{2}\omega)\}}$ in \cite{polson2013bayesian}, and $\tilde{f}_{i,n}^c=\sqrt{\mathbb{E}[{f_{i,n}^c}^2]}$. 

\textbf{Derivation for \cref{mf-f}}: The sigmoid transformation of latent functions, i.e., likelihood of the classification task, can be reformulated in the form of Gaussian distribution using P\'{o}lya-Gamma variables to ensure conjugation. Consequently, \cref{mf-f} is expressed as follows: 
\begin{equation}
\begin{aligned}
    &q_2(\mathbf{f})\\
    \propto& \prod_{i=1}^{T_r}\prod_{n=1}^{N_i^r}\mathcal{N}(f_{i,n}^r\mid y_{i,n}^r, \sigma_i^2)\prod_{i=1}^{T_c}\prod_{n=1}^{N_i^c}\mathcal{N}(f_{i,n}^c\mid \frac{y_{i,n}^c}{2\mathbb{E}[\omega_{i,n}]}, \frac{1}{\mathbb{E}[\omega_{i,n}]}) p(\mathbf{f})\\
    =&\prod_{i=1}^{T_r}\mathcal{N}(\mathbf{f}_i^r\mid \mathbf{y}_i^r, {\mathbf{D}_i^r}^{-1})\cdot\prod_{i=1}^{T_c}\mathcal{N}(\mathbf{f}_i^c\mid \frac{1}{2}{\mathbf{D}_i^c}^{-1}\mathbf{y}_i^c, {\mathbf{D}_i^c}^{-1})\cdot \mathcal{N}(0,\mathbf{K})\\
    =&\mathcal{N}(\bm{m},\mathbf{\Sigma}), 
\end{aligned}
\label{equ:mf-f}
\end{equation}
where $\mathbf{\Sigma}=(\mathbf{H}+\mathbf{K}^{-1})^{-1}$, $\bm{m}=\mathbf{\Sigma}\mathbf{H}\mathbf{v}$, $\mathbf{H}=\text{diag}(\mathbf{D_{\cdot}^{r},\mathbf{D}_{\cdot}^c})$ and $\mathbf{v}=[\mathbf{y}_{\cdot}^{r\top},\frac{1}{2}\mathbf{y}_{\cdot}^{c\top}\mathbf{D}_{\cdot}^{c-1}]^\top$ with $\mathbf{D}_i^r=\text{diag}(1/\sigma^2_i)$, $\mathbf{D}_i^c=\text{diag}(\mathbb{E}[\bm{\omega}_{i}])$.
\end{proof}

\section{Proof of Mean-field VI with Inducing Points}
\label{proof:mf_inducing}
\begin{proof}
Similar derivations have been provided in \cref{proof:mf}; here we restate the key formulas for clarity. The computation of \cref{equ:mf-f} involves inverting a matrix with cubic complexity, where the matrix size is determined by the sample size. 
In order to enhance the scalability of inference, $M$ inducing points $\mathbf{x}_1,\ldots,\mathbf{x}_M$ are randomly sampled from the existing dataset. The inducing outputs for the $i$-th task are denoted as $\mathbf{f}_{i,\mathbf{x}_m}$, and the collective inducing outputs across all tasks are denoted as $\mathbf{f}_{\mathbf{x}_m}=[\mathbf{f}^{\top}_{1,\mathbf{x}_m},\ldots,\mathbf{f}^{\top}_{T,\mathbf{x}_m}]^\top$ with the prior distribution $\mathbf{f}_{\mathbf{x}_m} \sim \mathcal{N}(\mathbf{0}, \mathbf{K}_{\mathbf{x}_m,\mathbf{x}_m})$. Specifically, the vector of function values for each task $\mathbf{f}_{i,\mathbf{x}_m}^{\cdot}\sim\mathcal{N}(\mathbf{0}, \mathbf{K}_{\mathbf{x}_m,\mathbf{x}_m}^{\cdot,i})$, where $\mathbf{K}_{\mathbf{x}_m,\mathbf{x}_m}^{\cdot,i}$ is the diagonal block of $\mathbf{K}_{\mathbf{x}_m, \mathbf{x}_m}$ corresponding to the specific task. 
For different tasks, we select the same inducing points to simplify the calculation of $\mathbf{K}_{\mathbf{x}_m, \mathbf{x}_m}$, as suggested by \cite{moreno2018heterogeneous}. 

Upon the introduction of inducing points, the likelihoods of the regression and classification tasks in \cref{joint-dist} are derived as follows, 
\begin{subequations}
\begin{equation}
    p(\mathbf{y}_i^r\mid \mathbf{f}_{i,\mathbf{x}_m}^{r}) = \int p(\mathbf{y}_i^r\mid \mathbf{f}_i^r)p(\mathbf{f}_i^r\mid\mathbf{f}_{i,\mathbf{x}_m}^{r})d\mathbf{f}_i^r,
    \label{inducing conditional:1}   
\end{equation}
\begin{equation}
    p(\mathbf{y}_i^c, \bm{\omega}_i\mid \mathbf{f}_{i,\mathbf{x}_m}^c)=\int p(\mathbf{y}_i^c, \bm{\omega}_i^c\mid \mathbf{f}_i^c)p(\mathbf{f}_i^c\mid \mathbf{f}_{i,\mathbf{x}_m}^c)d\mathbf{f}_i^c.
    \label{inducing conditional:2}
\end{equation}
\end{subequations}
Following the approach outlined in \cite{zhou2023heterogeneous}, we replace the distribution of data points $\mathbf{f}_i^{\cdot}$ conditional on inducing points $\mathbf{f}_{i, \mathbf{x}_m}^{\cdot}$ with a deterministic function to simplify computations. 
Specifically, we assume $\mathbf{f}_{i,\mathbf{x}_n}^{\cdot}$, the latent functions on predictive points $\mathbf{x}_n$, are the mean of $p(\mathbf{f}_{i,\mathbf{x}_n}^{\cdot}\mid \mathbf{f}_{i,\mathbf{x}_m}^\cdot)$: 
\begin{equation}
\mathbf{f}_{i,\mathbf{x}_n}^{\cdot}=\mathbf{K}_{\mathbf{x}_m,\mathbf{x}_n}^{\cdot, i^{\top}}\mathbf{K}_{\mathbf{x}_m,\mathbf{x}_m}^{\cdot, i^{-1}}\mathbf{f}_{i,\mathbf{x}_m}^{\cdot}, 
\label{simplified}
\end{equation}
where $\mathbf{K}_{\mathbf{x}_m,\mathbf{x}_n}^{\cdot, i}$ is the kernel w.r.t inducing points and predictive points. 

Substituting \cref{simplified} into \cref{equ:mf-f}, the optimal variational distribution of inducing outputs is derived as: 
\begin{equation}
\begin{aligned}
q(\mathbf{f}_{\mathbf{x}_m})
=&\prod_{i=1}^{T_r} \mathcal{N}(\mathbf{K}_{\mathbf{x}_m,\mathbf{x}_n}^{r, i\top}\mathbf{K}_{\mathbf{x}_m,\mathbf{x}_m}^{r, i^{-1}}\mathbf{f}_{i,\mathbf{x}_m}^{r}\mid \mathbf{y}_i^r, {\mathbf{D}_i^r}^{-1})\cdot \mathcal{N}(\mathbf{0}, \mathbf{K}_{\mathbf{x}_m,\mathbf{x}_m})\\
&\cdot\prod_{i=1}^{T_c}\mathcal{N}(\mathbf{K}_{\mathbf{x}_m,\mathbf{x}_n}^{c, i\top}\mathbf{K}_{\mathbf{x}_m,\mathbf{x}_m}^{c, i^{-1}}\mathbf{f}_{i,\mathbf{x}_m}^{c}\mid \frac{1}{2}{\mathbf{D}_i^{c}}^{-1}\mathbf{y}_i^c, {\mathbf{D}_i^c}^{-1})\\
= &\mathcal{N}(\mathbf{f}_{\mathbf{x}_m}\mid\mathbf{m}_{\mathbf{x}_m},\mathbf{\Sigma}_{\mathbf{x}_m}),
\end{aligned}
\end{equation}
$\mathbf{m}_{\mathbf{x}_m}=\mathbf{\Sigma}_{\mathbf{x}_m}[\mathbf{v}_{\mathbf{x}_m}^{r\top},\mathbf{v}_{\mathbf{x}_m}^{c\top}]^\top$, $\mathbf{\Sigma}_{\mathbf{x}_m}=\left[\text{diag}\left(\mathbf{H}_{\mathbf{x}_m}^r, \mathbf{H}_{\mathbf{x}_m}^c\right) + \mathbf{K}_{\mathbf{x}_m,\mathbf{x}_m}^{-1}\right]^{-1}$, $\mathbf{H}_{\mathbf{x}_m}^\cdot=\text{diag}(\mathbf{H}_{1,\mathbf{x}_m}^\cdot,\ldots,\mathbf{H}_{T_\cdot,\mathbf{x}_m}^\cdot)$, $\mathbf{v}_{\mathbf{x}_m}^\cdot=[\mathbf{v}_{1,\mathbf{x}_m}^{\cdot\top},\ldots,\mathbf{v}_{T_\cdot,\mathbf{x}_m}^{\cdot\top}]^\top$ and 
\begin{equation*}
\begin{gathered}
\mathbf{H}_{i,\mathbf{x}_m}^r=\mathbf{K}_{\mathbf{x}_m,\mathbf{x}_m}^{r,i^{-1}}\mathbf{K}_{\mathbf{x}_m,\mathbf{x}_n}^{r,i}\mathbf{D}^r_i\mathbf{K}_{\mathbf{x}_m,\mathbf{x}_n}^{r,i\top}\mathbf{K}_{\mathbf{x}_m,\mathbf{x}_m}^{r,i^{-1}},
\mathbf{v}_{i,\mathbf{x}_m}^{r}=\mathbf{K}_{\mathbf{x}_m,\mathbf{x}_m}^{r,i^{-1}}\mathbf{K}_{\mathbf{x}_m,\mathbf{x}_n}^{r,i}\frac{\mathbf{y}^r_i}{\sigma_i^2},\\
\mathbf{H}_{i,\mathbf{x}_m}^c=\mathbf{K}_{\mathbf{x}_m,\mathbf{x}_m}^{c,i^{-1}}\mathbf{K}_{\mathbf{x}_m,\mathbf{x}_n}^{c,i}\mathbf{D}^c_i\mathbf{K}_{\mathbf{x}_m,\mathbf{x}_n}^{c,i\top}\mathbf{K}_{\mathbf{x}_m,\mathbf{x}_m}^{c,i^{-1}},
\mathbf{v}_{i,\mathbf{x}_m}^{c}=\mathbf{K}_{\mathbf{x}_m,\mathbf{x}_m}^{c,i^{-1}}\mathbf{K}_{\mathbf{x}_m,\mathbf{x}_n}^{c,i}\frac{\mathbf{y}^c_i}{2}.
\end{gathered}
\end{equation*}

For each iteration on the client side, the computational complexity is $O((TM)^3+NM^2)$, where $N$ is the total number of training samples, $T$ is the number of tasks, and $M$ is the number of inducing points on each task. The computational complexity is dominated by matrix inversion $O((TM)^3)$ and product $O(NM^2)$. Given the assumption that $TM$ significantly smaller than $N$, the complexity can be simplified to $O(NM^2)$. 
\end{proof}
\section{Analytical Solution to ELBO}
\label{analytic elbo}
\begin{proof}
The calculation of ELBOs follows the same process for each client, hence we only derive the analytic solution of $\text{ELBO}_z$. The subscript $z$ is omitted in following statement, i.e. $\text{ELBO}$ hereafter.

\begin{equation}
\begin{aligned}
\text{ELBO}(\bm{\Theta})=&\underbrace{\mathbb{E}_{q_2(\{\mathbf{f}_i^r\}_{i=1}^{T_r})}[\log p(\mathbf{y}^{r}_{\cdot}\mid \{\mathbf{f}_{i}^r\}_{i=1}^{T_r})]}_{\text{(a)}}\\
&+\underbrace{\mathbb{E}_{q_1(\bm{\omega}),q_2(\{\mathbf{f}_i^c\}_{i=1}^{T_c})}[\log p(\mathbf{y}^{c}_{\cdot}\mid\bm{\omega},\{\mathbf{f}_{i}^c\}_{i=1}^{T_c})]}_{\text{(b)}} \\
&-\underbrace{\text{KL}(q_1(\bm{\omega})\Vert p(\bm{\omega}))}_{\text{(c)}} - \underbrace{\text{KL}(q_2(\mathbf{f})\Vert p(\mathbf{f}))}_{\text{(d)}}, 
\end{aligned}
\label{elbo-r}
\end{equation}
where $q_1(\bm{\omega})$, $q_2(\mathbf{f}_i^{\cdot})$ are optimal distribution of mean-field VI derived by \cref{equ:mf_local_q}. 

The expressions for the expectations of the log likelihood terms pertaining to regression tasks and classification tasks, i.e., (a), (b), are derived by recognizing the Gaussian distribution structure inherent in both terms: 
\begin{subequations}
\begin{equation}
    (a)=\sum_{i=1}^{T_r}\sum_{n=1}^{N_i^r}-\log(\sigma_i\sqrt{2\pi})-\frac{1}{2\sigma_i^2}({y_{i,n}^{r^2}}-2y_{i,n}^r\bar{f}_{i,n}^r+{\tilde{f}_{i,n}^{r^2}}), 
\end{equation}
\begin{equation}
(b)=\sum_{i=1}^{T_c}\sum_{n=1}^{N_i^c}\frac{y_{i,n}^c\bar{f}_{i,n}^c}{2}-\frac{{\tilde{f}_{i,n}^{c^2}}}{2}\mathbb{E}[\omega_{i,n}]-\log2. 
\end{equation}
\end{subequations}
where $\bar{f}_{i,n}^{\cdot}=\mathbb{E}[f_{i,n}^{\cdot}]$.

Moreover, the derivation of the KL divergence for P\'{o}lya-Gamma variables, i.e., (c), is accomplished through the general P\'{o}lya-Gamma distribution, $p_{\text{PG}}(\omega\mid b,c) = \frac{\exp(-\frac{c^2}{2}\omega)p_{\text{PG}}(\omega\mid b, 0)}{E_{\omega\sim p_{\text{PG}}(\omega \mid b,0)}\{\exp (-\frac{c^2}{2}\omega)\}}$, and Laplace transform, $\mathbb{E}_{\omega\sim p_{\text{PG}}(\omega\mid1,0)}\{\exp(-\omega t)\}=\frac{1}{\cosh(\sqrt{t/2})}$, in \cite{polson2013bayesian}: 
\begin{equation}
(c)=\sum_{i=1}^{T_c}\sum_{n=1}^{N_i^c}\log \cosh(\frac{\tilde{f}^c_{i,n}}{2})-\frac{\tilde{f}^c_{i,n}}{4}\tanh(\frac{\tilde{f}^c_{i,n}}{2}). 
\end{equation}

The derivation of the KL divergence for the latent function, i.e., (d), is the KL divergence of two Gaussian distributions, which has an analytical expression: 
\begin{equation}
\begin{split}
(d)=&\frac{1}{2}\big{(}\log\lvert\mathbf{K}\rvert-\log\lvert\mathbf{\Sigma}\rvert-N+\text{Tr}[\mathbf{K}^{-1}\mathbf{\Sigma}]+\mathbf{m}^{\top}\mathbf{K}^{-1}\mathbf{m}\big{)}. 
\end{split}
\end{equation}

Eventually, the application of \cref{equ:agg} for optimizing global prior, which is equivalent to optimization of hyperparameters $\bm{\Theta}=\big{\{}\{\bm{\phi}_b, \bm{\theta}_b\}_{b=1}^B, \mathbf{W}, \{\sigma_i^2\}_{i=1}^{T_r}\big{\}}$ is discussed below. For $\phi_b$, $\theta_b$ and $\mathbf{W}$, numerical optimization methods are employed to maximize the ELBO: 
\begin{subequations}
\begin{equation}
    \bm{\phi}_b^{(t+1)}=\bm{\phi}_b^{(t)}+\text{learning\_rate}\times \frac{\partial \frac{1}{\mathcal{Z}}\sum_{z=1}^{\mathcal{Z}}\text{ELBO}_z}{\partial\bm{\phi}_b}\big|_{\bm{\phi}_b^{(t)}}, 
\end{equation}
\begin{equation}
    \bm{\theta}_b^{(t+1)} = \bm{\theta}_b^{(t)}+\text{learning\_rate}\times \frac{\partial \frac{1}{\mathcal{Z}}\sum_{z=1}^{\mathcal{Z}}\text{ELBO}_z}{\partial \bm{\theta}_b}\big|_{\bm{\theta}_b^{(t)}}, 
\end{equation}
\begin{equation}
    \bm{\mathbf{W}}^{(t+1)}=\bm{\mathbf{W}}^{(t)}+\text{learning\_rate}\times\frac{\partial \frac{1}{\mathcal{Z}}\sum_{z=1}^{\mathcal{Z}}\text{ELBO}_z}{\partial \mathbf{W}}\big|_{\mathbf{W}^{(t)}}. 
\end{equation}
\end{subequations}
The choice of an efficient and stable numerical optimizer with an appropriately tuned learning rate is crucial, and we opt for the use of AdamW within the PyTorch framework.
    
For the regression noise $\{\sigma_i^2\}_{i=1}^{T_r}$, a closed-form expression for the optimal result can be obtained by observing that only term (a) in \cref{elbo-r} involves $\sigma^2_i$: 
\begin{equation}
    \hat{\sigma}^2_i=\frac{1}{N_i^r}\sum_{n=1}^{N_i^r}y_{i,n}^{r^2}-2y_{i,n}^r \bar{f}_{i,n}^r+\tilde{f}_{i,n}^{r^2}. 
    \label{optimial sigma}
\end{equation}
\end{proof}

\section{Solution to Multi-class Classification}
\label{multi_class}
In our paper, we confine the scope of classification tasks handled by our model to binary classification. Binary classification can be effectively modeled using a single latent function. And the augmented likelihood with P\'{o}lya-Gamma variables results in an analytical solution. However, it is more challenging for multi-class classification, a common scenario in real-world datasets. For a $K$-class classification task, the usual likelihood is a categorical distribution with softmax: 
$
    p(y_{i,n}^c=z\mid \{f_{i,n}^{c,k}\}_{k=1}^{K})=\frac{e^{f_{i,n}^{c,z}}}{\sum_{k=1}^K e^{f_{i,n}^{c,k}}},
$
where $f_{i,n}^{c,1},...f_{i,n}^{c,K}$ are $K$ latent functions on the input. 
However, the P\'{o}lya-Gamma augmentation technique can not be employed directly in the multi-class setting. 

To address this issue, many works have proposed corresponding solutions. 
Previous solutions include logistic-softmax function~\cite{galy2020multi,ke2023revisiting} and the one-vs-each softmax approximation~\cite{snell2020bayesian}. 
Both methods involve modifying the softmax-based likelihood into a new form, allowing the introduction of auxiliary latent variables using P\'{o}lya-Gamma augmentation. Through this way, the non-conjugate models are turned into conditional conjugate models. 
Both of these techniques can be seamlessly integrated into the framework we propose. We did not provide specific derivations here as they are beyond the scope of this paper. For details on these methods, please refer to~\cite{galy2020multi,ke2023revisiting,snell2020bayesian}. 

\section{Details of Experiment}
\label{app.experiment}

\subsection{Base Kernels in Ablation Study}
We compare the MOGP models with linear kernel, Laplace kernel, Cauchy kernel and RBF kernel, with expressions as follows: 
\begin{equation*}
    \begin{aligned}
        & \text{Linear Kernel: } k(\mathbf{x}, \mathbf{x}^\prime) = \mathbf{x}^\top\mathbf{x}^\prime,\\
        & \text{Laplace Kernel: } k(\mathbf{x}, \mathbf{x}^\prime) = \phi_0 \exp(-\frac{\phi_1}{2}\lvert\lvert \mathbf{x} - \mathbf{x}^\prime\rvert\rvert_1),\\
        & \text{Cauchy Kernel: } k(\mathbf{x}, \mathbf{x}^\prime) = \frac{1}{\phi_1 \lvert\lvert \mathbf{x} - \mathbf{x}^\prime \rvert\rvert^2_2 + 1},\\
        & \text{RBF Kernel: } k(\mathbf{x}, \mathbf{x}^\prime) = \phi_0 \exp(-\frac{\phi_1}{2}\lvert\lvert \mathbf{x} - \mathbf{x}^\prime\rvert\rvert_2^2), 
    \end{aligned}
\end{equation*}
where we set $\phi_0$, $\phi_1$ as $1$, $0.01$. It is worth noting that the inputs of the linear kernel are normalized by the L2-norm to ensure numerical stability. 

\end{document}